\documentclass[letterpaper, 10 pt, conference, onecolumn]{ieeeconf}  

\IEEEoverridecommandlockouts                              

\usepackage{float}
\usepackage{subfigure}

\usepackage{graphicx}
\graphicspath{{./images/}}
\usepackage{amssymb,amsmath}
\usepackage{bm} 
\usepackage{stackrel}
\usepackage{mathtools}

\DeclarePairedDelimiter{\ceil}{\lceil}{\rceil}
\DeclareMathOperator*{\argmax}{arg\,max}
\usepackage{xcolor}
\newtheorem{remark}{Remark}
\newtheorem{definition}{Definition}
\newtheorem{lemma}{Lemma}
\newtheorem{theorem}{Theorem}
\newtheorem{corollary}{Corollary}
\newtheorem{assumption}{Assumption}
\newtheorem{proofoflemma}{Proof of Lemma}
\newtheorem{proofoftheorem}{Proof of Theorem}
\title{\LARGE \bf A Decentralized Communication Policy for Multi Agent Multi Armed Bandit Problems}

\author{P.\,Pankayaraj$^{1}$, D. H. S. Maithripala$^{2}$
\thanks{$^{1}$Office of Research and Innovation Services, Sri Lanka Technological Campus, Padukka, CO 10500, Sri Lanka
        {\tt\small pankayarajp@sltc.ac.lk}}%
\thanks{$^{2}$ Department of Mechanical Engineering, University of Peradeniya, KY 20400, Sri Lanka / Sri Lanka Technological Campus, Padukka, CO 10500, Sri Lanka 
        {\tt\small smaithri@eng.pdn.ac.lk}}%
}

\begin{document}

\maketitle
\thispagestyle{empty}
\pagestyle{empty}

\allowdisplaybreaks

\begin{abstract}
This paper proposes a novel policy for a group of agents to, individually as well as collectively, solve a multi armed bandit (MAB) problem. The policy relies solely on the information that an agent has obtained through sampling of the options on its own and through communication with neighbors.   
The option selection policy is based on an Upper Confidence Based (UCB) strategy while the  communication strategy that is proposed forces agents to communicate with other agents who they believe are most likely to be exploring than exploiting. The overall strategy is shown to significantly outperform an independent Erd\H{o}s-R\'{e}nyi (ER) graph based random communication policy. 
The policy is shown to be cost effective in terms of communication and thus to be easily scalable to a large network of agents.
\end{abstract}

\section{Introduction} \label{sect:Introduction}
A slot machine with multiple levers, where the pulling of each leaver results in an unknown reward, is known as a \textit{Multi-Armed Bandit} (MAB). Each lever in the machine represents an option that the player, also referred to as an agent, has to make. 
The reward associated with each option is assumed to be the outcome of an independent stochastic process.  The agent is faced with the task of playing the machine in such a way that the cumulative reward it obtains is maximized.
This is known to be equivalent to minimizing the expected cumulative regret \cite{LaiRobbins}.  In their landmark work Lai and Robbins \cite{LaiRobbins} showed that  the cumulative regret was bounded below by a logarithmic function of the number of times the agent samples the options.  
They also devised a sampling rule that would guarantee logarithmically bounded cumulative regret. These results were refined initially in \cite{AgrawalSimpl} by introducing a confidence bound method and later extended to a family of methods known as Upper Confidence Bound (UCB) algorithms for ensuring asymptotic and uniform logarithmic cumulative regret. All these algorithms are based on a two pronged approach of first estimating the reward characteristics of the options, based on the results of the previous choices, and then executing a sampling strategy that in turn depends on the estimated rewards \cite{LaiRobbins,AgrawalSimpl,Auer,Kauffman}. The sampling strategy is typically based on an objective function that codifies an appropriate notion of balancing exploration and exploitation.
A common feature of these schemes is that they all ensure certain Hoeffding type asymptotic bounds on the tail probabilities of the estimates. It is known that the standard sample mean estimator also satisfies such tail bounds. This result holds true even if the reward distributions are non-stationary \cite{Garivier2011B}.

In MAB problems the upper bound on the total regret scales linearly with the number of options. Thus when the number of options become very large it becomes imperative that multiple agents be used to reduce the regret. When a group of agents collectively solve a MAB problem it will be referred to as a multi agent multi armed bandit (MAMAB) problem. Central to the problem is the appropriate use of communication. When the policy employed by an agent depends only on the information obtained by the agent, through sampling the options on its own and through communication with its neighbors, the problem will be referred to as a decentralized multi agent multi armed bandit (D-MAMAB) problem. The work by \cite{Kalathil,LandgrenECC,Kolla,LandgrenCDC,Patrick} extend the MAB problem to the decentralized multi-agent setting.
A running consensus, with agents observing the estimates of other agents through communication, is employed in \cite{LandgrenECC,LandgrenCDC,Patrick}. Fixed graphs as well as stationary stochastic communication graphs are considered in these studies.
A brief review of the existing D-MAMAB schemes that use a running consensus of the estimates is provided in \cite{Patrick}. They also propose a novel running consensus based algorithm that results in a lower regret than the other existing algorithms. The work by \cite{Kolla} considers a D-MAMAB scheme where the agents use a fixed communication graph to communicate only the instantaneous rewards obtained by the agents. This ideas is extended to the case of an independently and identically distributed Erd\H{o}s-R\'enyi (ER) graph based stochastic communication policy in \cite{LandgrenCDC,UdariNaomi}. 
To the best of our knowledge there exists no studies that employ a communication policy that depends on the past choices the agents have made. Such a communication policy will result in a non stationary communication graph.
We employ a UCB option allocation policy that is guaranteed to yield a logarithmically bounded regret even if the communicating graph is non-stationary and depends on the past choices the agents have made. The proposed option selection policy utilizes a reward estimate that depends on both the option rewards obtained by the agent itself as well as those obtained through communication. Redundancies in choice, due to two or more neighbors selecting the same option, are disregarded in the estimation. This estimate in a sense is the best estimate that an agent can make of the expectation of the option rewards solely based on the information it has obtained. 

Two types of regret are considered: one where each agent only considers the options that the agent samples itself, and another where agents consider only the reward values obtained through communication.  The regret that an agent accumulates by sampling suboptimal options by itself will be referred to as the \emph{self regret} while the regret that results due to the communication of the information of suboptimal  arms being picked by one of its neighbors will be referred to as the \emph{communication regret}. Redundancies occurring when more than one agent picks the same option are disregarded in the computation of the communication regret. 
When there is no communication or when the neighbors are picking the optimal option the communication regret will be zero. Thus the communication regret indicates if an agent is receiving information about suboptimal options without having to sample them itself. Thus the communication regret
serves as an indication of the \emph{effectiveness of communication}. The paper proposes, an option selection policy that maximizes an agent's self regret and a communication policy that increases the communication effect.
  
The paper proves that the proposed option selection policy guarantees that the self regret and the communication regret are both logarithmically bounded even if the communication graph depends on the past choices made by the individual agents. To the best of our knowledge it is the first time that such a result has been proven. 
The paper also shows that the self regret reduces with the increasing connectivity of the communication graph and that it then increases after a certain critical high connectivity value. The converse effect is demonstrated for the communication regret indicating increased effect of communication with increasing connectivity.  

In this paper we also propose a novel UCB based communication strategy to reduce the self regret and increase the communication regret.  In this communication policy agents communicate with other agents who they believe to be are exploring. Simulation results show that the policy only requires the communication with a few neighbors and that it significantly out performs an Erd\H{o}s-R\'{e}nyi (ER) graph based random communication strategy.
From a practical point of view, since the policy involves only the communication of two pieces of information and the policy depends only on local information, the scheme proposed here is easily scalable in a communication-ally and computationally cost effective manner. 

Section \ref{Secn:MAMABProblem} defines the D-MAMAB problem that is considered in this paper. The novel UCB based option selection policy and the proof that the policy guarantees logarithmic regret even if the communication graph is a dependent random graph is then presented in section \ref{Secn:RegretAnalysis}. Section 
\ref{Secn:UCBAgentPolicy} presents the proposed novel UCB based communication strategy. Finally the effectiveness of the proposed scheme is demonstrated through simulations in section \ref{Secn:Simulations}.

\section{The Decentralized Multi Agent Multi Armed Bandit Problem}\label{Secn:MAMABProblem}
Let the total number of agents be denoted by $n_A$ and the total number of options be denoted by $n_O$. The following explicit assumption states the class of Multi Armed Bandit Problems that will be considered in this work.

\begin{assumption}\label{as:MABassumption}
The reward associated with each option $i\in\{1,2,\cdots,n_O\}$ corresponds to a possibly non stationary stochastic process $\{X_i^t\}$ that satisfies the condition
$E\left(e^{\lambda X_i^t}\right)\leq e^{\lambda E(X_i^t)+\frac{\lambda^2{d_i}^2}{8}}$
for some $d_i>0$ and all $\lambda>0$. Furthermore it is also assumed that there exists $i_*\in\{1,2,\cdots,n_O\}$ and $\Delta, \bar{\Delta}>0$ such that $\Delta\leq E({X}_{i_*}^r)-E(X_{i}^s)\leq \bar{\Delta}$ for all $r,s>0$ and $i\in\{1,2,\cdots,n_O\}$ that satisfy $i \neq i_{*}$. 
\end{assumption}
\noindent This condition restricts the option rewards to be sub Gaussian  processes. Thus in particular the results derived here are also valid for any stochastic process that takes values in a  bounded interval of length $d_i$. The last statement of the above assumption also implies that there exists a well defined optimal option $i_*$. 

At each time instant $t$ each agent, based on the estimates of the expectations of each of the option rewards the agent has, chooses an option and chooses a set of agents to communicate with. It then updates its estimates of the expected values of the option rewards based on the reward obtained by sampling on its own as well based on the information obtained through communication. It then repeats the process in the next time step. At the initial time step $t=1$ agents sample the options based on some prior belief they may have of the option rewards. The strategy is decentralized as the policy only depends on the information that an agent obtains by itself as well through local communications.

The random variable $\varphi_j^t$, that takes values in $\{1,2,\cdots,n_O\}$, will denote the  option chosen by the $j^\mathrm{th}$ agent at time $t$. It will depend solely on the information available to the agent at time $t-1$.
Denote by $\mathcal{N}_{j}^t$ the set of agents who at time $t$ have communicated with $j$. It will be referred to as the \emph{neighbors of agent $j$ at time $t$}. This is the outcome of suitable policy that also depends solely on the information available to the agent at time $t-1$. Since the neighbors of agent $j$ denoted by the set $\mathcal{N}_j^t$ is a random variable the resultant communication graph $\mathcal{G}^t=(\mathcal{V},\mathcal{E}^t)$ is stochastic. 
By convention we let $j\in \mathcal{N}_{j}^t$. Let $\mathcal{N}_{j\alpha}$ be a subset of $\left\{1,2,\cdots,n_A\right\}$ that contains $j$ and let 
$\mathcal{N}_{jP}$ be the space of all such subsets of $\left\{1,2,\cdots,n_A\right\}$. The discrete random variable $\mathcal{N}_j^t$ takes values in $\mathcal{N}_{jP}$. We will denote the expectation of a random variable $f(\mathcal{N}_j^t)$ over the probability space of the graph by
$\left\langle f(\mathcal{N}_j^t)\right\rangle\triangleq  \sum_{\mathcal{N}_{j\alpha}\in \mathcal{N}_{jP}}f(\mathcal{N}_{j\alpha})\mathcal{P}\left( \mathcal{N}_j^t=\mathcal{N}_{j\alpha}\right)$.
This expectation is a constant if the stochastic communicating graph process $\{\mathcal{G}^t\}$ is an i.i.d. process.
We will also need the conditional expectation of $ f(\mathcal{N}_j^t)$ that is defined by
$\left\langle  f(\mathcal{N}_j^t)\right\rangle_{\mathcal{P}\left(\mathcal{N}_j^t\,|\,\varphi_{k}^t \right)} 
\triangleq\sum_{\mathcal{N}_{j\alpha}\in \mathcal{N}_{jP}}\mathcal{P}\left(\{\mathcal{N}_j^t=\mathcal{N}_{j\alpha}\}\,|\,\{\varphi_{k}^t=i\} \right)f(\mathcal{N}_{j\alpha})$.

\emph{The only information that $j$ receives from $k$ is the option that $k$ has chosen, $\varphi_k^t$, and the corresponding reward, $X_{\varphi_k^t}^t$, at that time instance.}  
We will denote the indicator random variable corresponding to the event $\mathcal{M}$ by $ \mathbb{I}_{\mathcal{M}}$. Then the Bernoulli random variable
$\mathbb{I}_{\{\varphi_j^t=i\}}$ denotes if the option $i$ was picked by agent $j$ at time $t$. The \emph{self reward} obtained by agent $j$ by sampling option $i$ is then given by
\begin{align}
S^s_{ij}(T)&=\sum_{t=1}^TX_i^t\mathbb{I}_{\{\varphi_{j}^t=i\}}.
\label{eq:SelfReward_ijT}
\end{align}


Let
$\mathcal{F}_j^{t}$ be the sigma algebra generated by the random variables 
\begin{align*}
\left\{\left(\mathcal{N}_j^\nu,\left\{X_{\varphi_r^\nu}^\nu\right\}_{r\in\mathcal{N}_j^\nu}\right)\right\}_{\nu=1}^{t}
\end{align*}
 and $\mathcal{F}_j^1\subset \mathcal{F}_j^2\subset \cdots \mathcal{F}_j^{t}$ be the corresponding filtration. Then $(\varphi_j^t,\mathcal{N}_j^t)$ are $\mathcal{F}_j^{t-1}$ measurable random variables.
The $\mathcal{F}_j^{t-1}$ measurable Bernoulli random variable 
\begin{align*}
\epsilon_{ij}^t\triangleq  \left\{
	\begin{array}{cl} 1 & \:\:\mathrm{if}\:\:\:\left(\sum_{k\in \mathcal{N}_j^t}\mathbb{I}_{\{\varphi_k^t=i\}}\right)\neq 0\\
  	0 & \:\:\: {\mathrm{o.w.}}\end{array}\right.,
\end{align*}
indicates whether $j$ has chosen option $i$ or one of its neighbors have chosen it at time $t$. 
Thus the $\mathcal{F}_j^{t-1}$ measurable random variable
\begin{align}
N_{ij}(t)&\triangleq  \sum_{\nu=1}^t\epsilon_{ij}^\nu,\label{eq:NijT}
\end{align}
denotes the total number of times that $j$ has become aware of $i$ being chosen in the time horizon $[1,2,\cdots,t]$.
The Bernoulli random variable $\mathbb{I}_{\{\varphi_{j}^t\neq i\:\&\:\exists\, k\,\in\mathcal{N}_{j}^t\,:\, \varphi_{k}^t=i\}}$ indicates whether or not $j$ has become aware of option $i$ being chosen at time $t$ when $j$ itself has not chosen option $i$. Then 
\begin{align}
S^c_{ij}(T)&\triangleq \sum_{i=1}^{n_O}\sum_{t=1}^TX_i^t\mathbb{I}_{\{\varphi_{j}^t\neq i\:\&\:\epsilon_{ij}^t=1\}}
\label{eq:UnConditionalCumulativeReward_ijT}
\end{align}
represents the reward that an agent obtains through communication and hence will be defined as the \emph{communication reward}. 

In the standard single agent MAB problem maximizing the reward is equivalent to minimizing the regret. Each time an agent adds the reward of an suboptimal option to its collected reward it also accumulates a regret that is equal to the difference between the optimal reward and the chosen suboptimal reward. Thus corresponding to the two types of rewards we define the \emph{self regret}, $R^s_{ij}(T)$ and the \emph{communication regret}, $R^c_{ij}(T)$ as follows:

{
\begin{align*}
R^s_{ij}(T)&={E\left(\sum_{t=1}^T\left(X_{i_*}^t-X_i^t\right)\mathbb{I}_{\{\varphi_{j}^t=i\}}\right)}\leq \bar{\Delta}{E\left(\sum_{t=1}^T\mathbb{I}_{\{\varphi_{j}^t=i\}}\right)}\\
R^c_{ij}(T)&={E\left(\sum_{t=1}^T\left(X_{i_*}^t-X_i^t\right)\mathbb{I}_{\{\varphi_{j}^t\neq i\:\&\:\epsilon_{ij}^t=1\}}\right)}\leq \bar{\Delta}{E\left(\sum_{t=1}^T\mathbb{I}_{\{\varphi_{j}^t\neq i\:\&\:\epsilon_{ij}^t=1\}}\right)}.
\end{align*}
}

\begin{remark}
In the MAMAB setting one looks for maximizing the effect of communication. The above expression for the communication regret, $R_{ij}^c(T)$, shows that a higher communication regret indicates that agents are communicating more with other agents who are sampling suboptimal options when they are not. This is highly desirable since this implies that an agent relies more on communication for exploration purposes. Thus one of the objectives of the MAMAB problem is to maximize the communication regret through the appropriate use of communication.
\end{remark}

\section{UCB Based Policy for MAMAB Problems}\label{Secn:RegretAnalysis}
The best possible estimate that an agent can make of the conditional expectation of $X_r^t$ is to use the full information it has access to. This optimal estimate is provided by $\widehat{X}_r^t$ that is defined by
\begin{align}
\widehat{X}_{ij}^t&\triangleq 
\frac{1}{N_{ij}(t)}\left(\sum_{\tau=1}^tX_i^\tau\epsilon_{ij}^\tau\right).\label{eq:SampleMean}
\end{align}
 
 A Hoeffding type tail bound is provided in Theorem 4 of \cite{Garivier2011B} for the random summand of pre-visible random variables. Setting $X_t=X_i^t$, $Y(t)=Y_{ij}(t)\triangleq \sum_{t=1}^T\left(X_i^t-E\left(X_i^t\right)\right)\epsilon_{ij}^t$, $\epsilon_t=\epsilon_{ij}^t$, and $N(t)=\sum_{\tau=1}^t\epsilon_{ij}^\tau$ in this result it follows that the above sample mean estimator satisfies the following tail bounds.
\begin{lemma}\label{lem:SampleMean}
The sample mean reward estimate $\widehat{X}_{ij}^t$ given by (\ref{eq:SampleMean}) satisfies the tail bound 
$\mathcal{P}\left(\left\{\left|\widehat{X}_{ij}^t-\widehat{\mu}_{ij}^t\right|>\sqrt{\frac{\Psi_j(t)}{N_{ij}(t)}}\right\}\right)\leq  
\frac{2}{t^2 \log (1+\eta)}$,
where $\Psi_j(t)$ satisfies $ 1.5({d_i}^2\sqrt{1+\eta})\log{(t)}\leq \Psi_{j}(t)$ for all $t>0$ and  $\eta>0$. Here $N_{ij}(t)$ is defined by (\ref{eq:NijT}) and $\widehat{\mu}_{ij}^t$ is given by
\begin{align}
\widehat{\mu}_{ij}^t&\triangleq 
\frac{1}{N_{ij}(t)}\left(\sum_{\tau=1}^tE\left(X_i^\tau\right)\epsilon_{ij}^\tau\right).\label{eq:ConditionalSampleMean}
\end{align}
\end{lemma}
When $\{X_i^t\}$ is a stationary processes the random variable $\widehat{\mu}_{ij}^t$ above reduces to the  expectation $\mu_{i}=E(X_i^t)$.
\begin{definition}
Let 	
\begin{align}
	Q_{sk}^t&\triangleq \widehat{X}_{sk}^t+\sqrt{\frac{\Psi_k(t)}{N_{sk}(t)}},\label{eq:UCBQ}
	\end{align}
	where $\widehat{X}_{sk}^t$ is defined by (\ref{eq:SampleMean}), $N_{sk}(t)$ is defined by (\ref{eq:NijT}), and  $\Psi_k(t)$ satisfies $ 1.5({d_s}^2\sqrt{1+\eta})\log{(t)}\leq \Psi_{k}(t)$ for all $t>0$ and some $\eta>0$.
The option selection policy $\{\varphi_{k}^{t}\}$ will be referred to as
\emph{UCB based}  if it is chosen such that
	\begin{align}
	\varphi_k^{t+1}=\argmax_{s}\limits\,\{Q_{sk}^t\}.\label{eq:UCBallocation}
	\end{align}
\end{definition}

The term $\Psi_k(t)$ in the above allocation rule dictates the exploration or the uncertainty of the estimates made by the agents. 
In the seminal paper by \cite{LaiRobbins} it is shown that for the MAB problem, the expected number of times that any suboptimal option is chosen by any optimal policy (that is given by $\sum_{\tau}^t\mathbb{I}_{\{\varphi_k^\tau=i\}}$), is necessarily bounded below by a logarithmic function of time. The work of \cite{Auer} and the extension by \cite{Garivier2011B} to include non-stationary bandits show that the sample mean estimator  (\ref{eq:SampleMean}), with  $\Psi_k(t)\sim \log{(t)}$, guarantees that a logarithmic bound is achieved. Crucial in the proof of this result is the observation that the probability of picking a suboptimal option when it has been picked more than a factor of $\Psi_k(t)$ is bounded by the tail probabilities of the estimator. This is stated formally in the lemma below and is proved in the appendix by using Lemma \ref{lem:SampleMean} and closely following the proof of \cite{LaiRobbins}. 

\begin{lemma}\label{Lemm:Main}
Let the conditions of assumption \ref{as:MABassumption} hold. 
Then any UCB based allocation rule $\{\varphi_k^t\}$ given by (\ref{eq:UCBQ}) -- (\ref{eq:UCBallocation}) will ensure that for all $i\neq i_*$
{
\begin{align*}
\mathcal{P}\left({\left\{\varphi_k^{t+1}=i\:\:\&\:\: N_{ik}(t)> \ceil*{\frac{4}{{\Delta^2}}\,\Psi_k(t)}\:\: \&\:\:{i \neq i_*}\right\}}\right)
\leq \frac{4}{t^2 \log (1+\eta)},
\end{align*}
}
\end{lemma}
\mbox{}\\

Using this Lemma and again by closely following \cite{LaiRobbins} it is shown in the appendix that the UCB based option allocation rule (\ref{eq:UCBallocation}) -- (\ref{eq:UCBQ}) guarantees that the self regret and the communication regret of an agent is bounded above as specified in the following theorem. 
\begin{theorem}\label{Theom:ijRegretDependent}
Let the conditions of assumption \ref{as:MABassumption} hold  and let $1.5({d_i}^2\sqrt{1+\eta})\log{t}\leq   \Psi_{k}(t) \leq  \Psi{(t)}$ for all $k\in\{1,2,\cdots,n_A\}$ and for some $\eta>0$.
Then the UCB based allocation rule $\{\varphi_j^t\}$ given by (\ref{eq:UCBQ}) -- (\ref{eq:UCBallocation}) will ensure that 
{
\begin{align*}
R^s_{ij}(T)&\leq  \bar{\Delta}\left(2+4\vartheta+f_i\left(\langle |\mathcal{N}_{j}^t|\rangle\right)\ceil*{\frac{4}{{\Delta^2}}\,\Psi(t)}\right),
\\
R_{ij}^c(T)&\leq  \bar{\Delta}\left(\max_{k,t\leq T}\left\langle |\mathcal{N}_{j}^t|\right\rangle_{\mathcal{P}\left(\mathcal{N}_j^t\,|\,\varphi_{k}^t \right)}-1\right)\, \left(2+4\vartheta+f_i\left(\langle |\mathcal{N}_{j}^t|\rangle\right)\ceil*{\frac{4}{{\Delta^2}}\,\Psi(t)}\right),
\end{align*}
}
for all $i\neq i_*$ and $j$ and $\vartheta=1/\log{(1+\eta)}$. Here $f_i\left(\langle |\mathcal{N}_{j}^t|\rangle\right)=\max_{k}f_{ik}\left(\langle |\mathcal{N}_{k}^t|\rangle\right)$ where
{
\begin{align}
f_{ik}\left(\langle |\mathcal{N}_{k}^t|\rangle\right)\!&\!\triangleq \frac{E\left(\sum_{t=2}^T\mathbb{I}_{\left\{\varphi_{k}^{t}=i \:\:\&\:\:N_{ik}(t-1)\leq \ceil*{\frac{4}{{\Delta^2}}\,\Psi(t-1)}\right\}} \right)}{ E\left(\sum_{t=1}^T\mathbb{I}_{\left\{\epsilon_{ik}^t=i \:\:\&\:\:N_{ik}(t-1)\leq \ceil*{\frac{4}{{\Delta^2}}\,\Psi(t-1)}\right\}} \right)}\!\leq \!\!1.\label{eq:fik}
\end{align}
}
\end{theorem}
This shows that when $\Psi(t)\sim \log{(t)}$ both the self regret and the communication regret are logarithmically bounded. 
\begin{remark}\label{rem:Fik}
Note that  $f_{ik}\left(\langle |\mathcal{N}_{k}^t|\rangle\right)=1$ when there is no communication between agents. As the expected connectivity $\langle |\mathcal{N}_{k}^t|\rangle$ increases $f_{ik}\left(\langle |\mathcal{N}_{k}^t|\rangle\right)$ reduce due to the fact that the denominator increases in comparison with the numerator. At the same time as the connectivity increases towards the maximum connectivity all agents have more or less the same information and thus they start behaving identically. Therefore the effect of communication may start to decrease with an associated increase in $f_{ik}\left(\langle |\mathcal{N}_{k}^t|\rangle\right)$.
Hence we conclude that the self regret will decrease as the connectivity increases and then start to increase as the connectivity approaches full connectivity. Simulation results show that this in fact is true. 
\end{remark}


In the special case where the communication graph, $\{\mathcal{G}^t\}$, is an i.i.d. process $\mathcal{P}\left(\mathcal{N}_j^t\,|\,\varphi_{k}^t \right)=\mathcal{P}\left(\mathcal{N}_j^t \right)$ is independent of time and thus $\max_{k,t}\left\langle |\mathcal{N}_{j}^t|\right\rangle_{\mathcal{P}\left(\mathcal{N}_j^t\,|\,\varphi_{k}^t \right)}=\langle |\mathcal{N}_j^t|\rangle$.
On the other hand if the connectivity of each node $j$ is restricted to $n_j$ then $\max_{k,t}\left\langle |\mathcal{N}_{j}^t|\right\rangle_{\mathcal{P}\left(\mathcal{N}_j^t\,|\,\varphi_{k}^t \right)}=n_j+1$. Thus we have the following corollary:
\begin{corollary}\label{Corro:ijRegretDependent}
Let the conditions of theorem \ref{Theom:ijRegretDependent} hold. Then if the communication graph is an i.i.d. stochastic process then the communication regret satisfies
{\small
\begin{align*}
R_{ij}^c(T)&\leq  \bar{\Delta}\left(\left\langle |\mathcal{N}_{j}^t|\right\rangle-1\right)\, \left(2+4\vartheta+f_i\left(\langle |\mathcal{N}_{j}^t|\rangle\right)\ceil*{\frac{4}{{\Delta^2}}\,\Psi(t)}\right),
\end{align*}
}
while if the graph is not an i.i.d. but the connectivity is restricted to $|\mathcal{N}_j^t|\leq n_j+1$ then the communication regret satisfies
{\small
\begin{align*}
R_{ij}^c(T)&\leq  \bar{\Delta}\,n_j\, \left(2+4\vartheta+f_i\left(\langle |\mathcal{N}_{j}^t|\rangle\right)\ceil*{\frac{4}{{\Delta^2}}\,\Psi(t)}\right).
\end{align*}
}
\end{corollary}

\section{UCB Based Communication Policy}\label{Secn:UCBAgentPolicy}
Theorem \ref{Theom:ijRegretDependent} shows that the UCB based option selection policy (\ref{eq:UCBQ}) -- (\ref{eq:UCBallocation})  guarantees the logarithmic bounding of the regret even if the communication graph depends on the past choices made by the agents.
In this section we propose such a novel communication policy to improve the performance of the regret. 

An agent choses its neighbors to communicate with based on a UCB type selection rule. The rule depends only on the information the agent has. The policy we propose encourages agents to communicate with other agents with whom they believe are most likely to be exploring. This is achieved by agents choosing to communicate with other agents based on \emph{who they estimate} to have the largest cost function values. We proceed to make this precise.

Let $\mathbb{I}_{\{j,k\}}^t$ denote the $\mathcal{F}_{t-1}$ measurable Bernoulli random variable that denotes if or not agent $j$ communicates with agent $k$. We do not require that the communication be bi-directed. That is we do not require that $\mathbb{I}_{\{j,k\}}^t=\mathbb{I}_{\{k,j\}}^t$. Each agent maintains an estimate of the rewards estimated by other agents using only the information made available through communication. That is, we let $\widehat{X}_{ijk}^t$ be the estimate that $j$ makes of the estimate that $k$ has made of the reward of the option $i$ within the time horizon $[1,2,\cdots,t]$. Explicitly stated
\begin{align}
\widehat{X}_{ijk}^t&\triangleq 
\frac{1}{N_{ijk}(T)}\left(\sum_{\nu=1}^TX_i^\nu\mathbb{I}_{\{j,k\}}^\nu\mathbb{I}_{\{\varphi_k^\nu=i\}}\right).\label{eq:MeanAverageijk}
\end{align}
Here
\begin{align}
	N_{ijk}(t)&\triangleq \sum_{\nu=1}^t\mathbb{I}_{\{j,k\}}^\nu\mathbb{I}_{\{\varphi_k^\nu=i\}},\label{eq:Nijkt}
\end{align}
is the random variable that denotes the number of times that $j$ has been made aware of by agent $k$ that it has chosen option $i$. 
The definition below makes precise the UCB based novel communication policy that we propose in this paper.
\begin{definition}\label{def:UCB basedcommunication}
Let $n_j$ be the maximum number of other agents that agent $j$ is allowed to communicate with and let
\begin{align}
	Q_{ijk}^t&\triangleq \widehat{X}_{ijk}^t+\sqrt{\frac{\Psi_{jk}(t)}{N_{ijk}(t)}}.\label{eq:UCBQc}
	\end{align} 
	where $\widehat{X}_{ijk}^t$ is defined by (\ref{eq:MeanAverageijk}), $N_{ijk}(t)$ is defined by (\ref{eq:Nijkt}), and  $\Psi_{jk}(t)$ satisfies $1.5({d_s}^2\sqrt{1+\eta})\log{(t)}\leq \Psi_{jk}(t)$ for all $t>0$ and some $\eta>0$.
Define $\mathcal{Q}_{j}^t$ to be the  largest $n_j$ values of the set $\cup_{k\neq j}\max \{Q_{ijk}^t\,|\,
\:\:i=1,2\cdots,n_O\: s.t.\: i\neq\varphi_j^t\}$. 
When ambiguity arises due to repeated elements the ambiguity will be resolved by choosing in a uniformly random manner. Then agent $j$ chooses its neighbors according to the policy
\begin{align}
\mathcal{N}_j^{t}=\left\{\arg_{k} Q_{ijk}^t \:\:| \:\: Q_{ijk}^t\in\mathcal{Q}_{j}^t\right\}\cup\{j\}.\label{eq:UCB_Agent_allocation}
\end{align}
	\end{definition}
	
	Note that (\ref{eq:UCB_Agent_allocation}) implies
	{\small
	\begin{align*}
	\mathbb{I}_{\{j,k\}}^t=\left\{
	\begin{array}{cl} 1 & \:\:\mathrm{if}\:\:\:Q_{ijk}^t\in\mathcal{Q}_{j}^t\\
  	0 & \:\:\: {\mathrm{o.w.}}\end{array}\right.
	\end{align*}}
\begin{remark}
This UCB based communication policy ensures that an agent communicates with other agents who the agent  believes to be most likely to be exploring than exploiting at that time instant.
\end{remark}

Theorem \ref{Theom:ijRegretDependent} guarantees that the resultant agent regrets remain logarithmically bounded if one uses the UCB based option allocation policy  (\ref{eq:UCBQ}) -- (\ref{eq:UCBallocation}). In the simulations below we show that this policy significantly outperforms an i.i.d. ER graph random communication policy.
Notice that since the communication involves only two pieces of locally gathered information the scheme is easily scalable and is very communication-ally and computationally cost effective. 

\section{Simulations}\label{Secn:Simulations}\label{Secn:Simulations}
In this section we use simulations to demonstrate the effectiveness of the UCB based option selection policy (\ref{eq:UCBQ}) -- (\ref{eq:UCBallocation})  and the UCB based communication policy (\ref{eq:UCB_Agent_allocation}). We compare the results with that of a random ER graph based communication policy and show that the proposed policy outperforms it significantly especially in the case of low connectivities.
We also demonstrate how the self regret decreases and the communication regret increases with increased connectivity indicating the effectiveness of communication.

\begin{figure}[H]
		\centering
		\includegraphics[scale = 0.6]{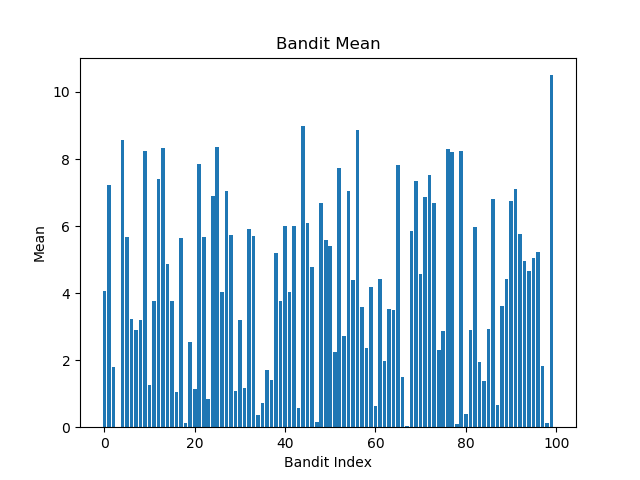}
		\caption{Actual Mean of the 5 bandit configuration  }
		\label{fig:Mean20}

\end{figure}

The number of bandits chosen for the simulation is 100 while the number of agents chosen is 20. Each bandit reward is assumed to satisfy a Gaussian normal process with variance equal to 2. The mean of those distributions were chosen as shown in figure \ref{fig:Mean20}. At the initial time step each agent $k$ initializes its  estimates, $\widehat{X}_{ik}^0$ by randomly sampling from a probability distribution that represents its prior belief of the option rewards. A time horizon of $T=20,000$ was chosen for each agent and the expectations were estimated by averaging over 4 trials.
In the case where the communication graph process $\mathcal{G}^t$ is an iid ER graph process, the graph is an element of the space of all possible ER graphs of $n_A$ nodes and edge probability $p$ ($\mathbb{G}(n_A,p)$) at each time $t$. In this case the probability of $j$ having the set of neighbors $\mathcal{N}_j^t=\mathcal{N}_{j\alpha}$ is given by
$ \mathcal{P}\left( \mathcal{N}_j^t=\mathcal{N}_{j\alpha}\right)=p^{|\mathcal{N}_{j\alpha}|-1}(1-p)^{n_A-|\mathcal{N}_{j\alpha}|}$.
Note that $\langle |\mathcal{N}_{j\alpha}|\rangle =(n_A-1)p+1$.

\begin{figure}[H]
	\centering	
	\includegraphics[scale = 0.4]{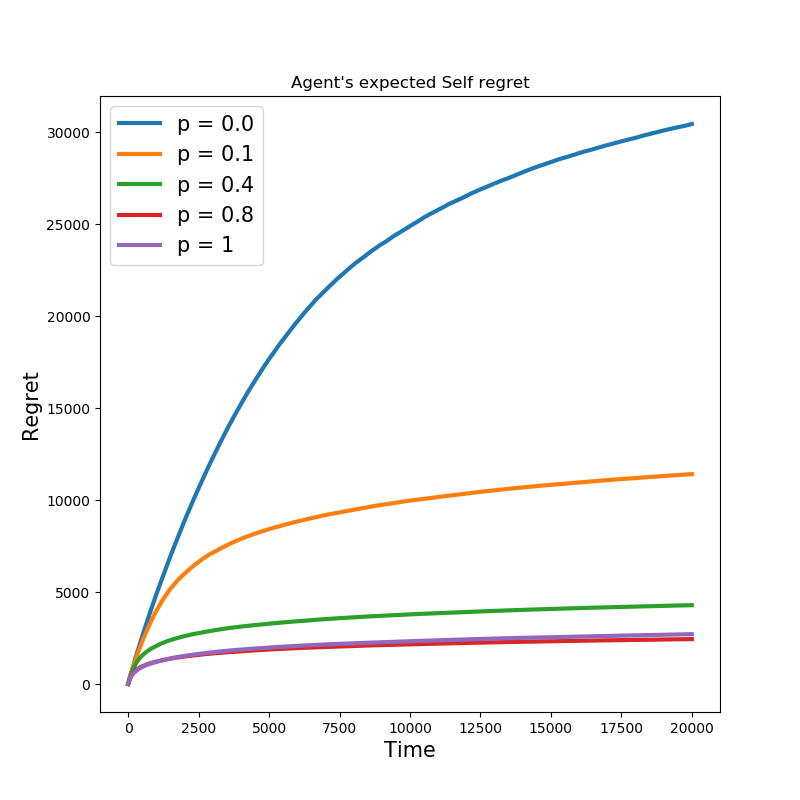}
	\includegraphics[scale = 0.4]{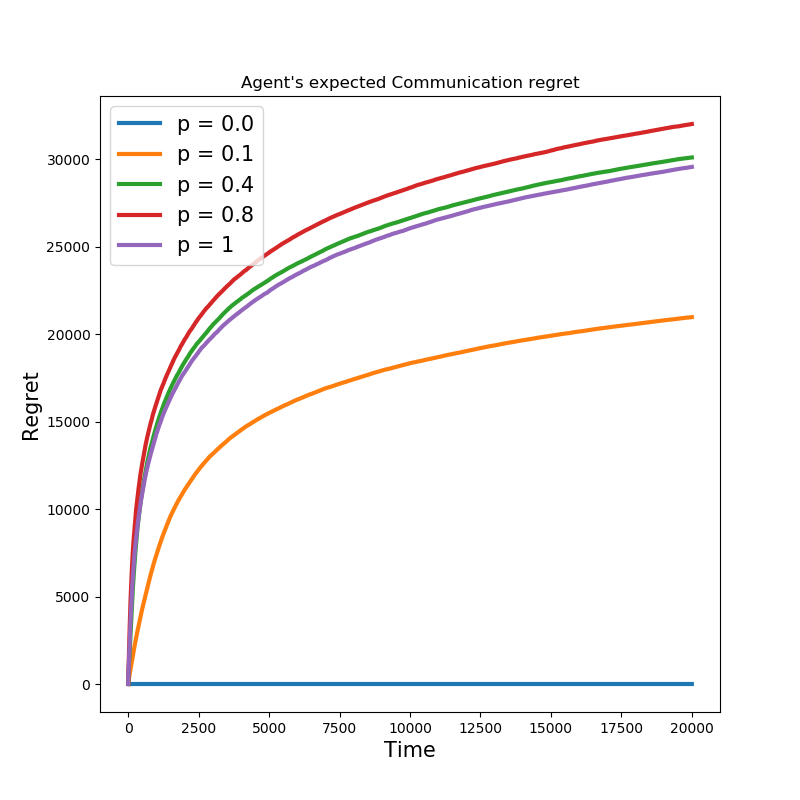}
	\caption{The expected values of the network self and communication regrets per agent ($\sum_{j}R^s_{ij}(T)/n_A, \sum_{j}R^c_{ij}(T)/n_A$) for several communication probabilities $p$ of the iid ER graph communication policy for 20 agents and  100 options.}	\label{fig:ERRegret}
	\end{figure}

\begin{figure}[H]
	\centering	
		\includegraphics[scale = 0.4]{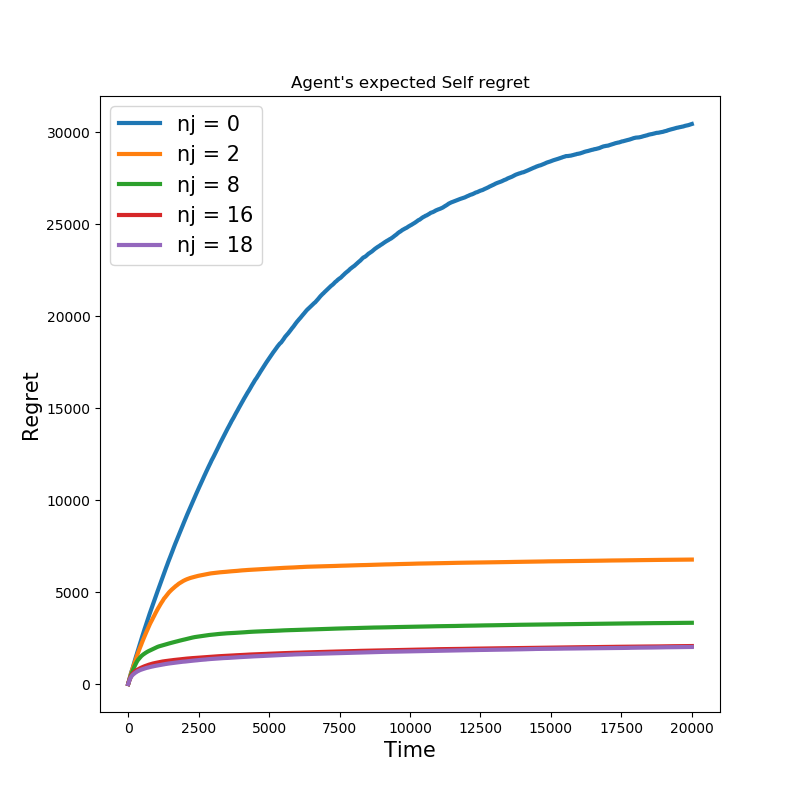}
		\includegraphics[scale = 0.4]{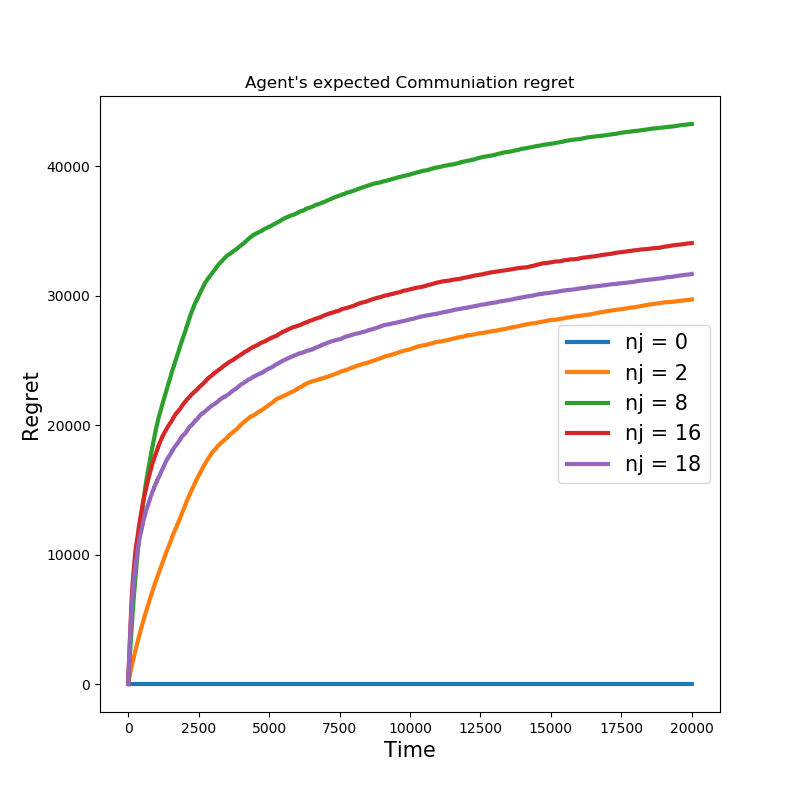}
		\caption{The expected values of the network self and communication regrets per agent ($\sum_{j}R^s_{ij}(T)/n_A, \sum_{j}R^c_{ij}(T)/n_A$)  for several fixed edge connectivities, $n_j$ for the UCB graph communication policy (\ref{eq:UCB_Agent_allocation}) for 20 agents and  100 options.}\label{fig:UCB_Regret}	
	\end{figure}

In what follows we compare the performance of the UCB based option selection policy (\ref{eq:UCBQ}) -- (\ref{eq:UCBallocation}) for different values of the connectivity of the graph for the two cases: a) when the communication is based on a random i.i.d. ER graph process and b) when the communication is based on the UCB based communication policy (\ref{eq:UCB_Agent_allocation}).
Figure \ref{fig:ERRegret} 
shows the estimates of the network regret per agent for several communication probabilities $p$ for the i.i.d. ER graph communication policy for the two cases of self and communication regret while figure \ref{fig:UCB_Regret} show the corresponding graphs when the 
communication is based on the UCB based communication policy (\ref{eq:UCB_Agent_allocation}).
Figures \ref{fig:ERRegret} and \ref{fig:UCB_Regret} clearly demonstrate that communication has a favorable effect on the self regret and as predicted by corollary \ref{Corro:ijRegretDependent}. It also show that increased connectivity increases the self regret indicating an improved communication effect.

\begin{figure}[H]
 \centering
	\includegraphics[scale = 0.4]{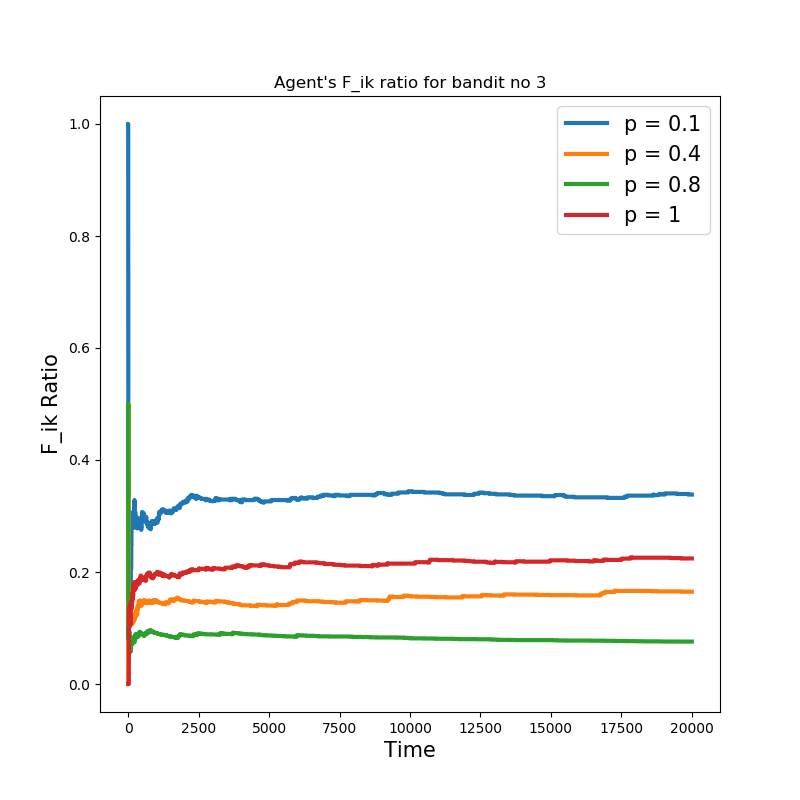}		
	\includegraphics[scale = 0.4]{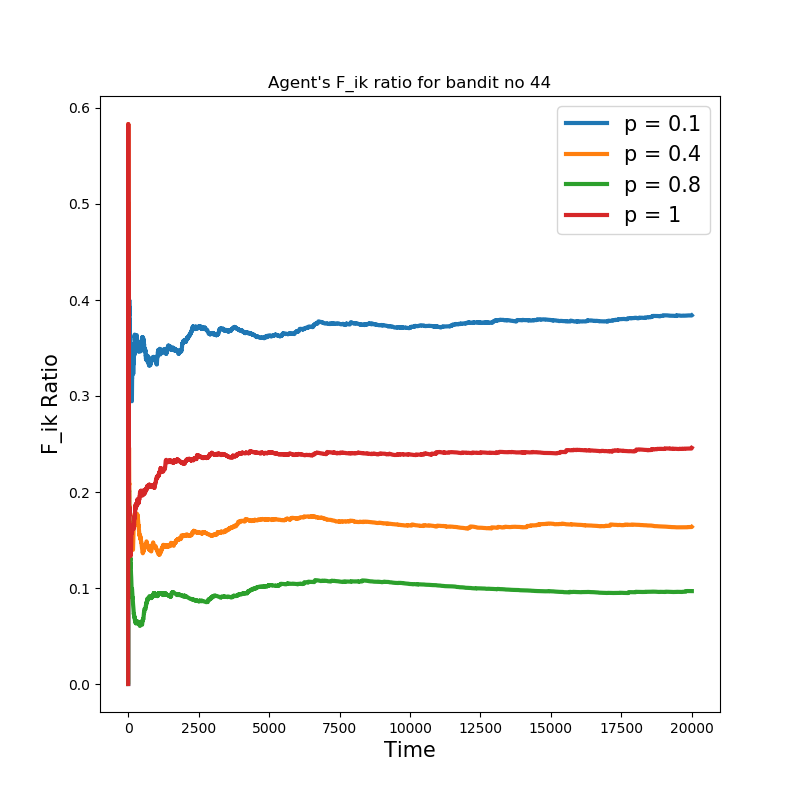}

	\caption{The fraction $f_{ij}(\langle\mathcal{N}_j^t\rangle)$, given by (\ref{eq:fik}), for agent 4 is plotted for the lowest and the highest  suboptimal option respectively for different connectivity values of the i.i.d. ER based communication policy.}\label{fig:fikER}
\end{figure}

\begin{figure}[h]
\centering
	\includegraphics[scale = 0.4]{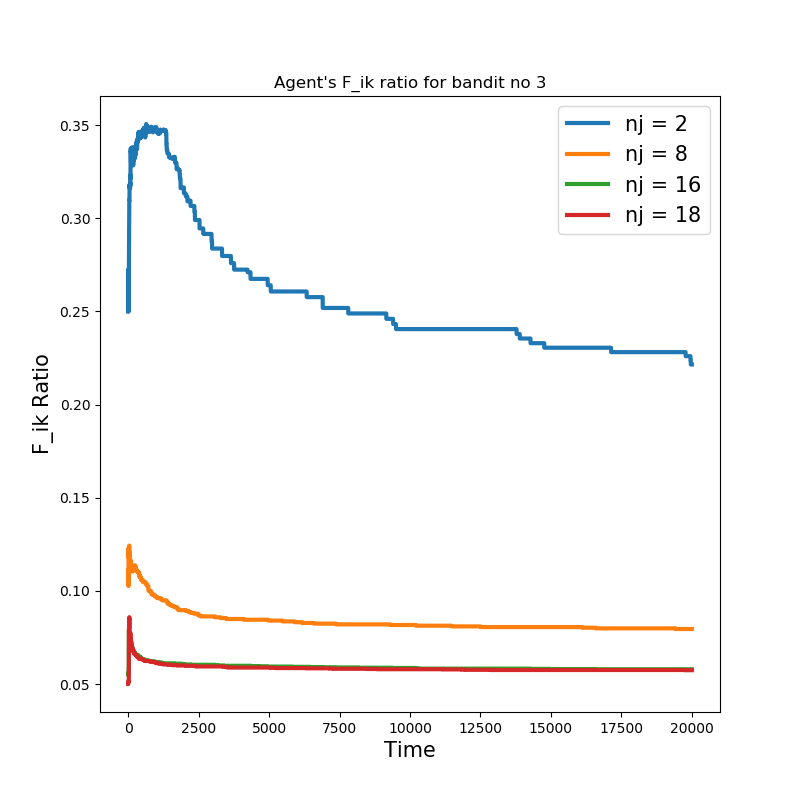}		
	\includegraphics[scale = 0.4]{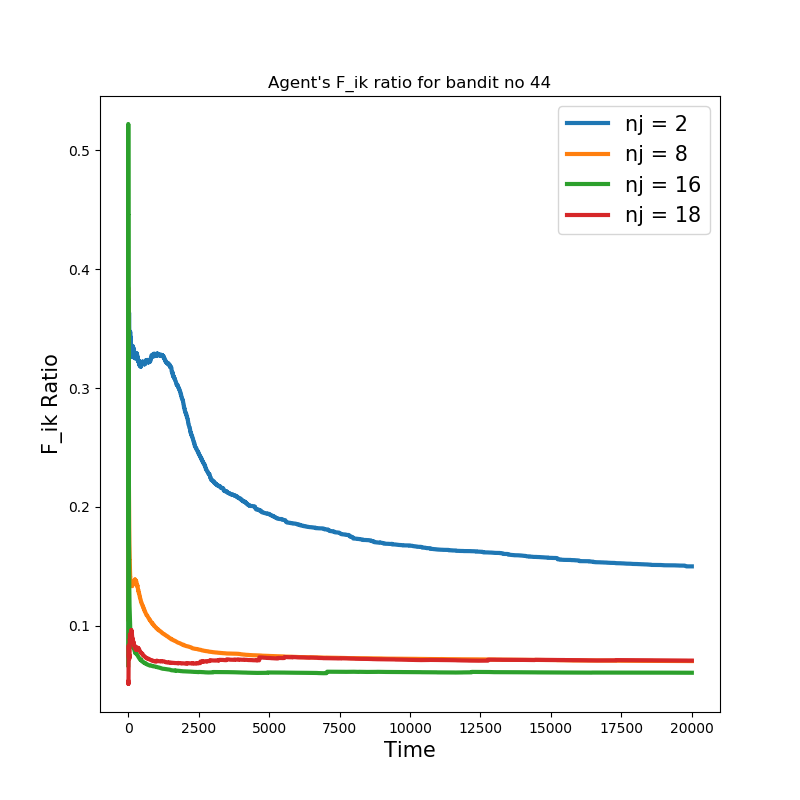}
	\caption{The fraction $f_{ij}(\langle\mathcal{N}_j^t\rangle)$, given by (\ref{eq:fik}), for agent 4 is plotted for the lowest and the highest suboptimal option respectively for different connectivity values of the UCB based communication policy (\ref{eq:UCB_Agent_allocation}).}\label{fig:fikUCB}
\end{figure}

\begin{figure}[H]
	\centering

\includegraphics[scale = 0.25]{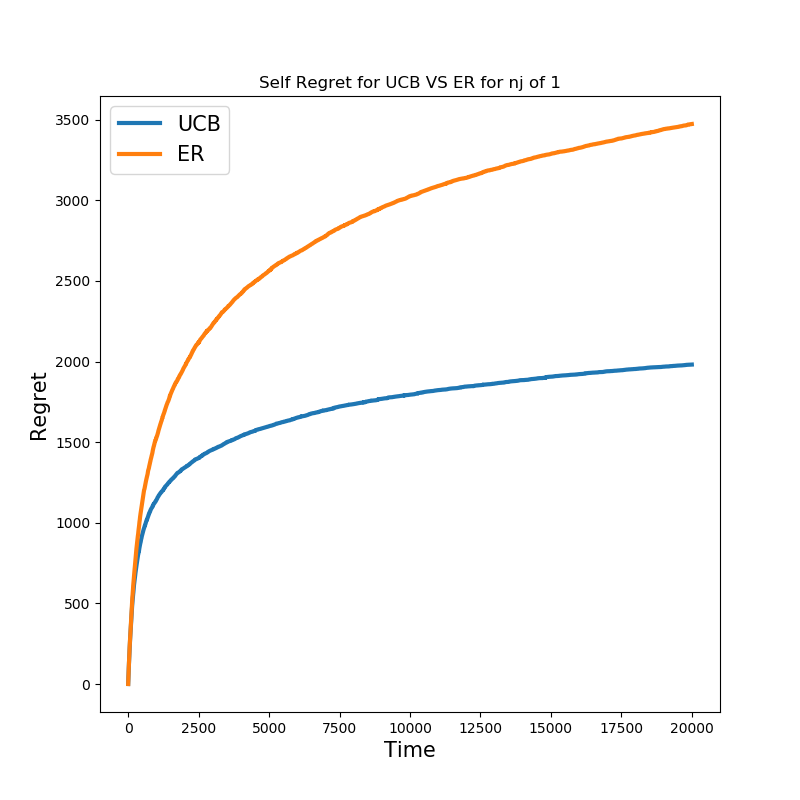}
	\includegraphics[scale = 0.25]{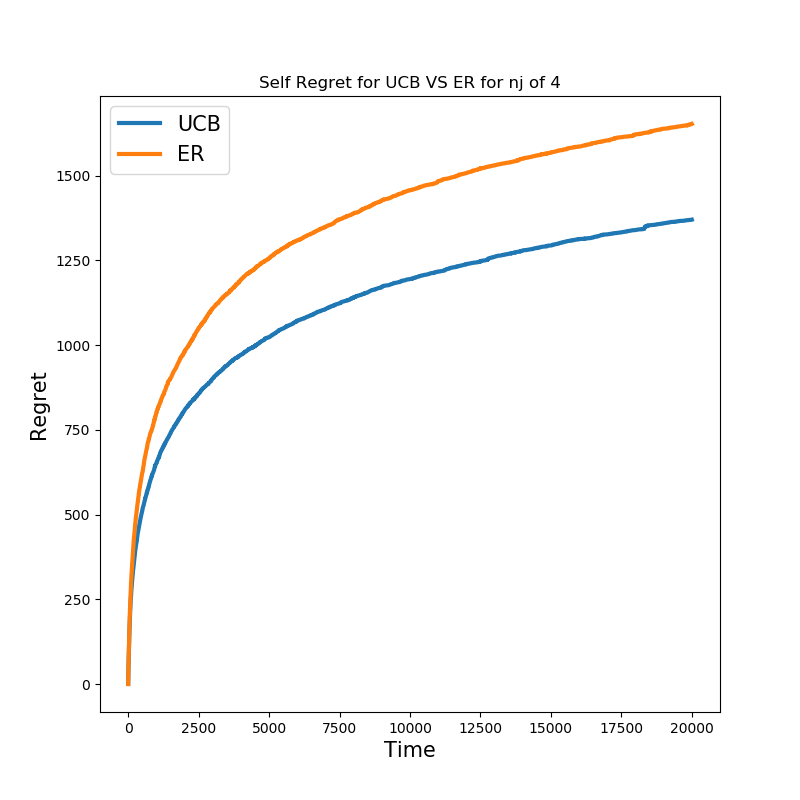}
	\includegraphics[scale = 0.25]{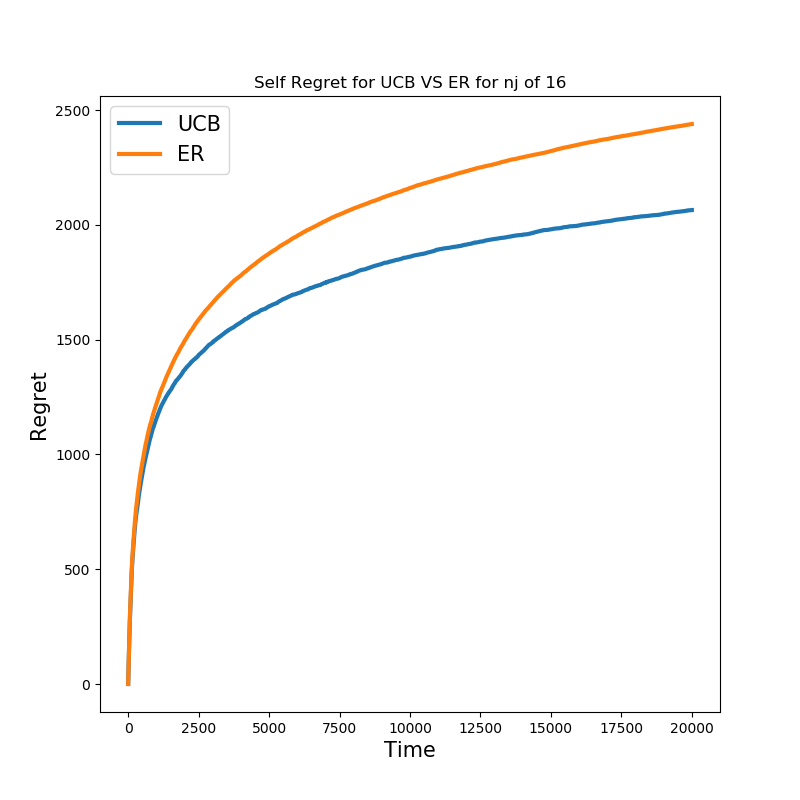}

	\caption{The comparison of the network self regret per agent, $\sum_{j}R_{ij}^s(T)/n_A$, between the ER graph communication policy vs the UCB based communication policy (\ref{eq:UCB_Agent_allocation}) for an expected connectivity of 2,4 and 16 respectively.
	}\label{fig:dependen1}
	
\end{figure}

\begin{figure}[H]
	\centering	
	\includegraphics[scale = 0.25]{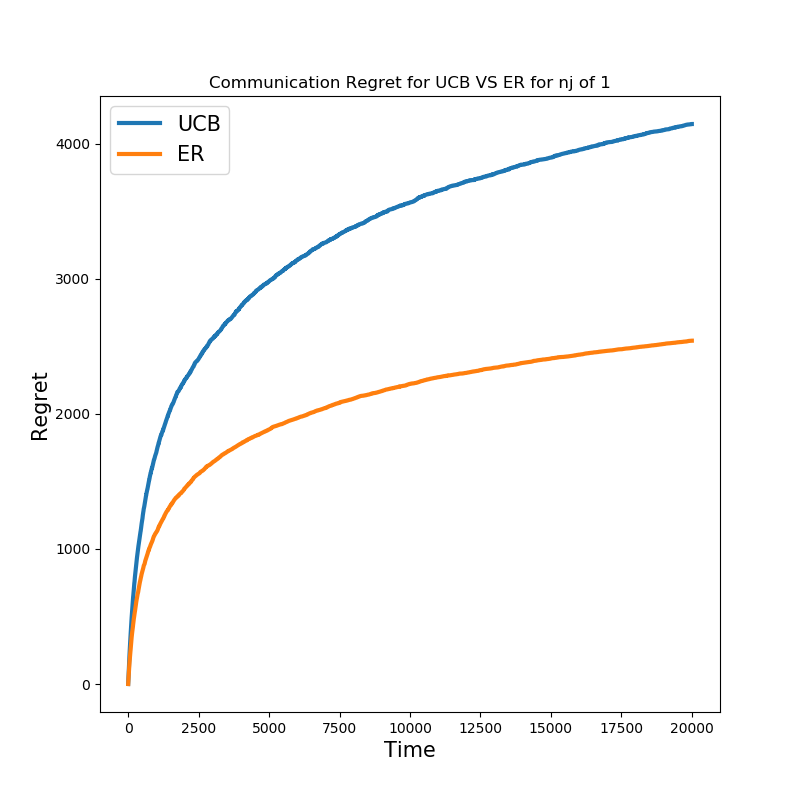}
	\includegraphics[scale = 0.25]{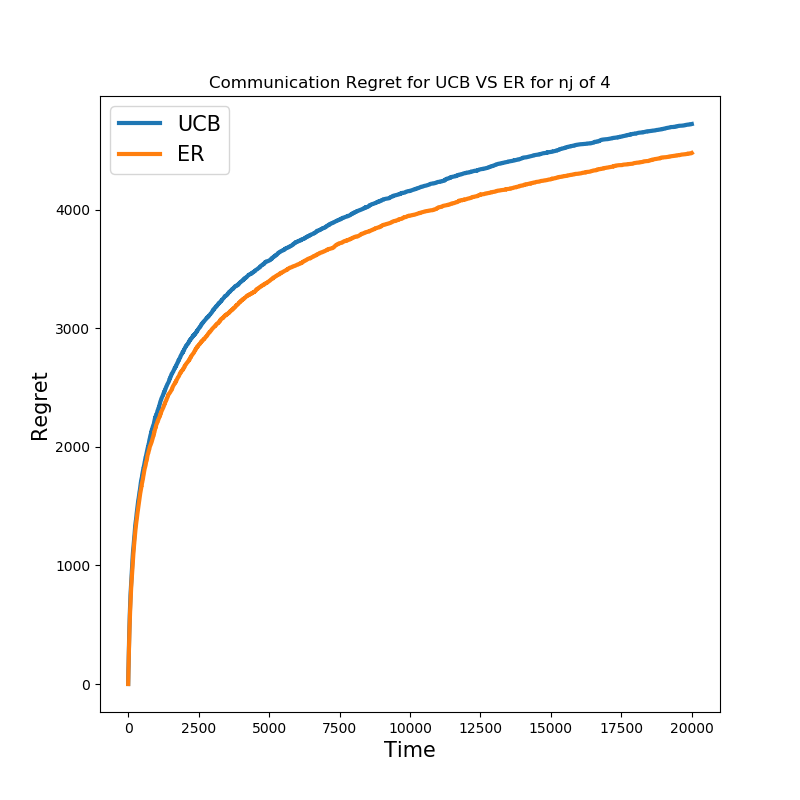}
	\includegraphics[scale = 0.25]{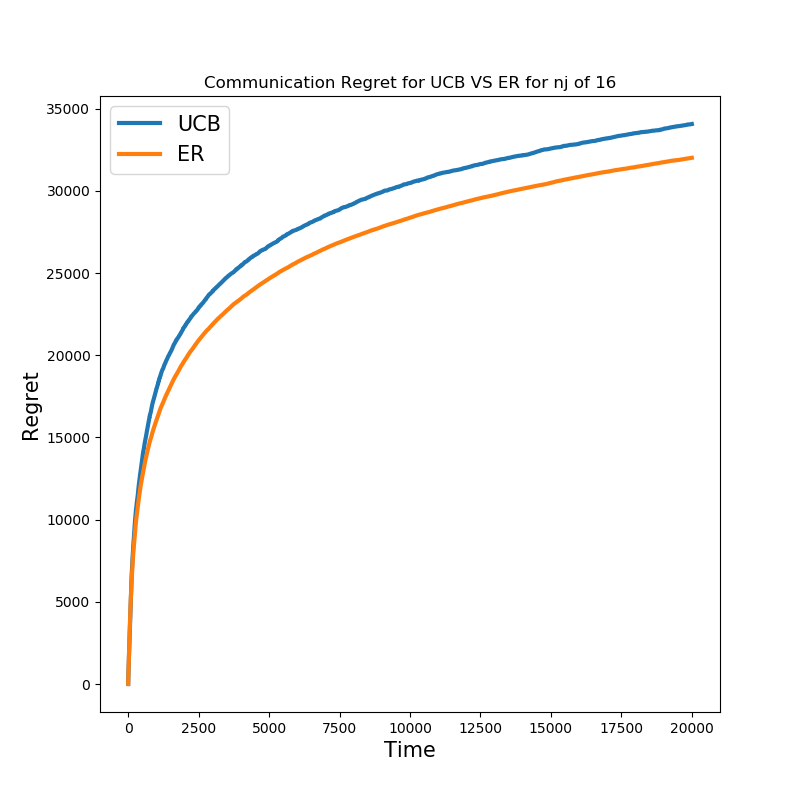}	
	
	\caption{The comparison of network communication regret per agent, $\sum_{j}R_{ij}^c(T)/n_A$, between the ER graph communication policy vs the UCB based communication policy (\ref{eq:UCB_Agent_allocation}) for an expected connectivity of 2,4 and 16 respectively. 
	}\label{fig:dependen2}
\end{figure}

The results also confirm the assertion of  remark \ref{rem:Fik} that the self regret initially decreases with increasing connectivity, and then increases beyond a certain optimal connectivity value while the inverse relationship is demonstrated for the communication regret. The effect is less noticeable for the UCB based communication policy. Repeating the experiment for different number of options showed no change in this qualitative behavior with the critical connectivity remaining the same. However it showed a change when the number of agents changed with the critical connectivity changing as the number of agents changed.
Comparing figures \ref{fig:fikER} -- \ref{fig:fikUCB} corresponding to the largest suboptimal option one also sees that the factor $f_{ij}(\langle\mathcal{N}_j^t\rangle)$ given by (\ref{eq:fik}) is much less for the UCB communication policy as opposed to the random ER graph based communication policy. This explains the superior performance of the former as demonstrated in figure \ref{fig:dependen1} and figure \ref{fig:dependen2} especially in the case of low connectivity. 
Figure \ref{fig:UCB_Regret} and figure \ref{fig:fikUCB} also show that, for the UCB based communication policy the increasing the connectivity beyond $n_j=8$ does not result in a noticeable change in the self regret. Thus indicating that a few connections are sufficient to achieve superior performance.


\section{Conclusion}
The paper reports a UCB based decentralized option selection policy and a novel UCB based communication policy for solving the MAMAB problem. The proposed option selection policy is shown to guarantee the logarithmic bounding of the regret irrespective of the communication policy that is employed. It is the first time that such a result has been proven. The paper also demonstrates that the UCB based communication strategy with a few local connections out performs a highly connected random ER graph based communication strategy. 



\bibliographystyle{IEEEtran}
\bibliography{DynamicBandit}



\begin{appendix}

\subsection{Logarithmic Regret Bounds}
\addtocounter{proofoflemma}{1}
\begin{proofoflemma}
In the following we will prove the Lemma \ref{Lemm:Main} by closely following the proof provided in \cite{LaiRobbins}. Let $C_{ik}(t)\triangleq  \sqrt{\frac{\Psi(t)}{N_{ik}(t)}}$. 

For any $i\neq i_{*}$ and some $0<l<t$ define
\begin{align*}
\mathcal{A}_{ik}^t&\triangleq\{\widehat{X}_{i_*k}(t)+C_{i_*k}(t)\geq \widehat{\mu}_{i_*k}(t)\},\\
\mathcal{B}_{ik}^t&\triangleq\{\widehat{\mu}_{i_*k}(t)> \widehat{\mu}_{ik}(t)+2{C_{ik}(t)}\},\\
\mathcal{C}_{ik}^t&\triangleq\{\widehat{\mu}_{ik}(t)+2{C_{ik}(t)}\geq \widehat{X}_{ik}^t+{C_{ik}(t)}\},
\end{align*}
where $\widehat{\mu}_{ik}(t)$ is the conditional expectation of the estimate $\widehat{X}_{ik}^t$.
Then we have,
\begin{align*}
\{\mathcal{A}_{ik}^t\cap \mathcal{B}_{ik}^t \cap \mathcal{C}_{ik}(t) \}\subseteq \{Q_{i_*k}^t> {Q_{ik}^t}\}.
\end{align*}
This implies that,
\begin{align*}
\{Q_{i_*k}^t\leq{Q^t_{ik}}\} \subseteq {\bar{\mathcal{A}}_{ik}^t}\bigcup {\bar{\mathcal{B}}_{ik}^t} \bigcup {\bar{\mathcal{C}}_{ik}^t},
\end{align*}
where the over bar denotes the complement of the set.
Thus for some $l>0$
\begin{align*}
{\{ Q_{i_*k}^t\leq{Q^t_{ik}}\:\:\&\:\: N_{ik}(t)> l\:\: \&\:\:{i \neq i_*}\}}\subseteq {\bar{\mathcal{A}}_{ik}^t}\bigcup {\widetilde{\mathcal{B}}_{ik}^t} \bigcup {\bar{\mathcal{C}}_{ik}^t},
\end{align*}
where
\begin{align*}
\widetilde{\mathcal{B}}_{ik}^t&\triangleq \bar{\mathcal{B}}_{ik}^t\cap {\{ Q_{i_*k}^t\leq{Q^t_{ik}}\:\:\&\:\: N_{ik}(t)> l\:\: \&\:\:{i \neq i_*}\}}.
\end{align*} 

From the above expressions we have,
{\small
\begin{align*}
\mathcal{P}({\{Q_{i_*k}^t\leq{Q^t_{ik}}\:\:\&\:\: N_{ik}(t)> l\:\: \&\:\:{i \neq i_*}\}})&\leq \mathcal{P}({\bar{\mathcal{A}}_{ik}^t})+\mathcal{P}({\widetilde{\mathcal{B}}_{ik}^t})+\mathcal{P}({\bar{\mathcal{C}}_{ik}^t}).
\end{align*}
}
and hence that
\begin{align*}
\mathcal{P}({\{Q_{i_*k}^t\leq{Q^t_{ik}}\:\:\&\:\: N_{ik}(t)> l\:\: \&\:\:{i \neq i_*}\}})&\leq \mathcal{P}(\widetilde{\mathcal{B}}^t_{ik})+2\max_{s}\mathcal{P}(\{|\widehat{X}_{sk}^t -\widehat{\mu}_{sk}(t)|>C_{sk}^t\}).
\end{align*}

What remains to complete the bound is to find an upper bound for $\widetilde{\mathcal{B}}^t_{ik}$. \emph{We consider the case where there exists a well defined optimal arm at all times. That is the case where there exists a $\Delta, \bar{\Delta}>0$ such that $\Delta\leq E(X_{i_*}^r)-E(X_i^s)\leq \bar{\Delta}$ for all $r,s>0$.}
Then 
since $\bar{\mathcal{B}}^t_{ik}=\{\widehat{\mu}_{i_*k}(t)\leq \widehat{\mu}_{ik}(t)+2{C^t_{ik}}\}$ the UCB based allocation rule (\ref{eq:UCBQ}) implies that,
\begin{align*}
\bar{\mathcal{B}}^t_{ik}&=
\left\{\frac{\Delta}{2}\leq \sqrt{\frac{\Psi(t)}{N_{ik}(t)}}\right\},
\end{align*}
where $\Psi(t)$ is such that $\Psi_{k}(t)\leq \Psi(t)$ for all $k$.

Thus since $\widetilde{\mathcal{B}}^t_{ik}\subseteq \bar{\mathcal{B}}_{ik}^t\cap \left\{Q_{i_*k}^t\leq{Q^t_{ik}}\:\:\&\:\: N_{ik}(t)> l\:\:\&\:\: i\neq i_*\right\}$
we have that
{
\begin{align*}
&\mathcal{P}(\widetilde{\mathcal{B}}^t_{ik})
\leq\mathcal{P}\left(\left\{N_{ik}(t)\leq \frac{4}{{\Delta^2}}\,\Psi(t)\:\:\&\:\: N_{ik}(t)> l\right\}\right).
\end{align*}
}

Since when 
\begin{align*}
l(t)&\triangleq\ceil*{\frac{4}{{\Delta^2}}\,\Psi(t)},
\end{align*}
\begin{align*}
\mathcal{P}(\widetilde{\mathcal{B}}^t_{ik})
\leq\mathcal{P}\left(\left\{N_{ik}(t)\leq \frac{4}{{\Delta^2}}\,\Psi(t)\:\:\&\:\: N_{ik}(t)>\, l(t)\right\}\right)=0,
\end{align*}
we have shown that
for any $i\neq i_*$
\begin{align*}
\mathcal{P}\left({\left\{Q_{i_*k}^t\leq{Q^t_{ik}}\:\:\&\:\: N_{ik}(t)> \ceil*{\frac{4}{{\Delta^2}}\,\Psi(t)}\:\: \&\:\:{i \neq i_*}\right\}}\right)\leq 2\max_{s}\mathcal{P}\left(\left\{\left|\widehat{X}_{sk}^t- \widehat{\mu}_{sk}(t)\right|>\sqrt{\frac{\Psi_{k}(t)}{N_{sk}(t)}}\right\}\right).
\end{align*}
Note that
\begin{align*}
\left\{\varphi_k^{t+1}=i\:\:\&\:\: N_{ik}(t)> \ceil*{\frac{4}{{\Delta^2}}\,\Psi(t)}\:\: \&\:\:{i \neq i_*}\right\}&
= \left\{Q_{i_*k}^t\leq{Q^t_{ik}}\:\:\&\:\: N_{ik}(t)> \ceil*{\frac{4}{{\Delta^2}}\,\Psi(t)}\:\: \&\:\:{i \neq i_*}\right\}
\end{align*}
and this along with Lemma \ref{lem:SampleMean} completes the proof of Lemma \ref{Lemm:Main}. 
\end{proofoflemma}



\begin{proofoftheorem}
We begin the regret analysis by noting that the self and communication regrets, of agent $j$ due to sampling of the non optimal arm $i$ satisfies
\begin{align*}
R_{ij}^s(T)&\leq \bar{\Delta}\sum_{t=1}^T\mathcal{P}\left(\{\varphi_{j}^t=i\}\right),\\
R_{ij}^c(T)&\leq 
\bar{\Delta}\sum_{t=1}^T\left(
\sum_{\mathcal{N}_{j\alpha}\in \mathcal{N}_{jP}}\mathcal{P}\left( \{\mathcal{N}_j^t=\mathcal{N}_{j\alpha}\}\right)\mathcal{P}\left(\{\varphi_{j}^t\neq i\:\&\:\exists k\in\mathcal{N}_{j\alpha}\,:\, \varphi_{k}^t=i\}\,|\, \{\mathcal{N}_j^t=\mathcal{N}_{j\alpha}\}\right)\right).
\end{align*}
Note that for $l(t)\triangleq \ceil*{\frac{4}{{\Delta^2}}\,\Psi(t)}$
\begin{align*}
\{\varphi_{k}^t=i\}=\{\varphi_{k}^t=i\:\:\&\:\:N_{ik}(t-1)\leq\,l(t-1)\}\cup\{\varphi_{k}^t=i\:\:\&\:\:N_{ik}(t-1)> l(t-1)\}.
\end{align*}
Thus  
\begin{align*}
\mathcal{P}(\{\varphi_{k}^t=i\})
&= \mathcal{P}(\{\varphi_{k}^t=i\:\:\&\:\:N_{ik}(t-1)\leq\,l(t-1)\})+\mathcal{P}(\{\varphi_{k}^t=i\:\:\&\:\:N_{ik}(t-1)> l(t-1)\}).
\end{align*}
Hence  from Lemma \ref{Lemm:Main} we have that for an efficient reward estimator 
\begin{align*}
\sum_{t=1}^T\mathcal{P}(\{\varphi_{k}^t=i\})&\leq\sum_{t=2}^T\mathcal{P}(\{\varphi_{k}^t=i\:\:\&\:\:N_{ik}(t-1)\leq\,l(t-1)\})+2+\sum_{t=2}^T\frac{2\vartheta}{(t-1)^2}\\
&\leq E\left(\sum_{t=2}^T\mathbb{I}_{\{\varphi_{j}^t=i\:\:\&\:\:N_{ij}(t-1)\leq l(t-1)\}}\right)+2+4\vartheta
\end{align*}

Hence we have
\begin{align*}
R^s_{ij}(T)&\leq \:\bar{\Delta}\left(E\left(\sum_{t=2}^T\mathbb{I}_{\{\varphi_{j}^t=i\:\:\&\:\:N_{ij}(t-1)\leq l(t-1)\}}\right)+2+4\vartheta\right).
\end{align*}

Let $\mathcal{N}_{jP}$ be the space of all subsets of $\left\{1,2,\cdots,n_A\right\}$ that contain $j$. Then we see that
\begin{align*}
\sum_{t=1}^T\mathcal{P}\left(\{\varphi_{j}^t\neq i\:\&\:\epsilon_{ij}^t=1\}\right)
&\leq \sum_{t=1}^T\mathcal{P}\left(\cup_{\mathcal{N}_{j\alpha}\in \mathcal{N}_{jP}}\cup_{\stackrel{k\in \mathcal{N}_{j\alpha}}{k\neq j}}\left\{\{\varphi_{k}^t=i\}\,\cap\,\{\mathcal{N}_{j}^t=\mathcal{N}_{j\alpha}\}\right\}\right)\\
&\leq \sum_{t=1}^T\sum_{\mathcal{N}_{j\alpha}\in \mathcal{N}_{jP}}\sum_{\stackrel{k\in \mathcal{N}_{j\alpha}}{k\neq j}}\mathcal{P}\left(\{\varphi_{k}^t=i\} \cap\{\mathcal{N}_j^t=\mathcal{N}_{j\alpha}\}\right)\\
&\leq \sum_{t=1}^T\sum_{\mathcal{N}_{j\alpha}\in \mathcal{N}_{jP}}\max_k\mathcal{P}\left(\{\mathcal{N}_j^t=\mathcal{N}_{j\alpha}\}\,|\,\{\varphi_{k}^t=i\} \right)\sum_{\stackrel{k\in \mathcal{N}_{j\alpha}}{k\neq j}}\mathcal{P}\left(\{\varphi_{k}^t=i\} \right)\\
&\leq  \sum_{\mathcal{N}_{j\alpha}\in \mathcal{N}_{jP}}\max_{k,t\leq T}\mathcal{P}\left(\{\mathcal{N}_j^t=\mathcal{N}_{j\alpha}\}\,|\,\{\varphi_{k}^t=i\} \right)\sum_{\stackrel{k\in \mathcal{N}_{j\alpha}}{k\neq j}}\sum_{t=1}^T\mathcal{P}\left(\{\varphi_{k}^t=i\} \right)\\
&\leq \left(\max_k\sum_{t=1}^T\mathcal{P}\left(\{\varphi_{k}^t=i\} \right)\right)\sum_{\mathcal{N}_{j\alpha}\in \mathcal{N}_{jP}}\max_{k,t\leq T}\mathcal{P}\left(\{\mathcal{N}_j^t=\mathcal{N}_{j\alpha}\}\,|\,\{\varphi_{k}^t=i\} \right)(|\mathcal{N}_{j\alpha}|-1)\\
&\leq \left(\max_k\sum_{t=1}^T\mathcal{P}\left(\{\varphi_{k}^t=i\} \right)\right)\left(\max_{k,t\leq T}\left\langle |\mathcal{N}_{j}^t|-1\right\rangle_{\mathcal{P}\left(\mathcal{N}_j^t\,|\,\varphi_{k}^t \right)}\right)
\end{align*}
where we have defined
\begin{align*}
\left\langle |\mathcal{N}_{j}^t|-1\right\rangle_{\mathcal{P}\left(\mathcal{N}_j^t\,|\,\varphi_{k}^t \right)} 
&\triangleq\sum_{\mathcal{N}_{j\alpha}\in \mathcal{N}_{jP}}\mathcal{P}\left(\{\mathcal{N}_j^t=\mathcal{N}_{j\alpha}\}\,|\,\{\varphi_{k}^t=i\} \right)(|\mathcal{N}_{j\alpha}|-1)
\end{align*}
Thus we have that the communication regret satisfies
\begin{align*}
R_{ij}^c(T)&\leq\bar{\Delta}\sum_{t=1}^T
\mathcal{P}\left(\{\varphi_{j}^t\neq i\:\&\:\epsilon_{ij}^t=1\}\right)\\
&\leq  \left(\max_{k,t\leq T}\left\langle |\mathcal{N}_{j}^t|-1\right\rangle_{\mathcal{P}\left(\mathcal{N}_j^t\,|\,\varphi_{k}^t \right)}\right)\,\bar{\Delta} \left(\max_{k}E\left(\sum_{t=2}^T\mathbb{I}_{\{\varphi_{k}^t=i \:\:\&\:\:N_{ik}(t-1)\leq l(t-1)\}} \right)+2+4\vartheta\right).
\end{align*}

It is clear that
\begin{align*}
E\left(\sum_{t=2}^T\mathbb{I}_{\{\varphi_{k}^t=i \:\:\&\:\:N_{ik}(t-1)\leq l(t-1)\}} \right)\leq E\left(\sum_{t=1}^T\mathbb{I}_{\{\epsilon_{ik}^t=i \:\:\&\:\:N_{ik}(t-1)\leq l(t-1)\}} \right)\leq l(T),
\end{align*}
with equality holding if the connectivity, $|\mathcal{N}_{t}^t|-1$, is zero.
Define
\begin{align*}
f_{ik}\left(\langle |\mathcal{N}_{k}^t|\rangle\right)&\triangleq \frac{E\left(\sum_{t=2}^T\mathbb{I}_{\{\varphi_{k}^t=i \:\:\&\:\:N_{ik}(t-1)\leq l(t-1)\}} \right)}{ E\left(\sum_{t=1}^T\mathbb{I}_{\{\epsilon_{ik}^t=i \:\:\&\:\:N_{ik}(t-1)\leq l(t-1)\}} \right)}.
\end{align*}
Then we have $E\left(\sum_{t=2}^T\mathbb{I}_{\{\varphi_{k}^t=i \:\:\&\:\:N_{ik}(t-1)\leq l(t-1)\}} \right)\leq l(t-1)f_{i}\left(\langle |\mathcal{N}_{j}^t|\rangle\right)$ where $f_{i}\left(\langle |\mathcal{N}_{j}^t|\rangle\right)=\max_{k}f_{ik}\left(\langle |\mathcal{N}_{k}^t|\rangle\right)$ and we have proved the theorem.
\end{proofoftheorem}

\end{appendix}



\end{document}